\documentclass[12pt]{article}
\usepackage{times}  
\usepackage{helvet} 
\usepackage{courier}  
\usepackage[hyphens]{url}  
\usepackage{graphicx} 
\usepackage{graphicx}  
\usepackage{natbib}

\newenvironment{sciabstract}{%
\begin{quote} \baselineskip14pt\small\hfil {\bf Abstract} \hfil\\[3pt]}
{\end{quote}\vspace{6pt}}

\newcounter{lastnote}

\usepackage{times}
\usepackage{soul}
\usepackage{url}
\usepackage[small]{caption}
\usepackage{graphicx}
\usepackage{amsmath}
\usepackage{amsthm}
\usepackage{booktabs}
\usepackage{algorithm}
\usepackage{algorithmic}
\usepackage{epsfig,subfigure,multirow,booktabs,bm,threeparttable,booktabs}

\title{Novelty-Prepared Few-Shot Classification}
\author
{Chao Wang$^{1,2}$, Ruo-Ze Liu$^{1}$, Han-Jia Ye$^{1}$, Yang Yu$^{1}$\\
\normalsize{$^1$National Key Laboratory for Novel Software Technology,}\\
\normalsize{Nanjing University, Nanjing 210023, China}\\
\normalsize{$^2$Polixir}\\
\normalsize{E-mail: chao.wang@polixir.ai}
}

\date{}

\begin{document}

\baselineskip16pt

\maketitle 

\begin{sciabstract}
Few-shot classification algorithms can alleviate the data scarceness issue, which is vital in many real-world problems, by adopting models pre-trained from abundant data in other domains.
However, the pre-training process was commonly unaware of the future adaptation to other concept classes.
We disclose that a classically fully trained feature extractor can leave little embedding space for unseen classes, which keeps the model from well-fitting the new classes.
In this work, we propose to use a novelty-prepared loss function, called self- compacting softmax loss (SSL), for few-shot classification.
The SSL can prevent the full occupancy of the embedding space.
Thus the model is more prepared to learn new classes.
In experiments on CUB-200-2011 and mini-ImageNet datasets, we show that SSL leads to significant improvement of the state-of-the-art performance.
This work may shed some light on considering the model capacity for few-shot classification tasks.
\end{sciabstract}

\section{Introduction}
The use of deep neural network (DNN) to solve visual recognition tasks has been a huge success~\citep{DBLP:conf/nips/KrizhevskySH12}.
However, the deep neural network relies heavily on a large number of labeled images from a set of pre-defined visual categories, which in contrast, humans can learn to recognize new classes from a few examples, even a single one~\citep{carey1978acquiring}.
The high cost of getting labeled examples limits the scalability and practicality of deep neural network.
Because of its challenging and practical prospects, few-shot learning~\citep{DBLP:conf/cogsci/LakeSGT11,DBLP:conf/cvpr/MillerMV00,DBLP:journals/pami/Fei-FeiFP06} has attracted widespread attention, which aims to recognize novel visual categories from very few examples.

A variety of few-shot learning methods have been proposed to tackle few-shot classification tasks.
These methods can be divided into two categories: Meta-learning based methods and Pre-training based methods.
However, the pretraining process used in the above methods was commonly unaware of the future adaptation to other concept classes.
We find that the embedding space obtained by a feature extractor trained using softmax loss (SL) is almost entirely occupied by seen classes, leaving little embedding space for unseen classes.
This reduces the generalization performance on few-shot classification tasks.
We argue that one way to solve the above problem is to add a compact optimization objective while maintaining the classification optimization objective of SL.
To this end, we propose a novelty-prepared loss function, called \textbf{Self-compacting Softmax Loss} (SSL).
By adjusting the classification decision boundaries between prototypes, SSL can prevent the full occupancy of the embedding space.
Thus the model obtained is more prepared to learn new classes.
Besides, existing methods pass the input through a series of high-to-low resolution sub-network blocks, get a low-resolution representation finally.
We argue that there is a loss of valid information in this process because the high-resolution features are a useful complement to few-shot classification.
Therefore, to retain more useful information of only limited data, we use the high-resolution network in few- shot classification tasks.

Experiments are conducted on two benchmark datasets to compare the proposed method and the related state-of-the-art methods for few-shot classification.
On CUB-200-2011 , the proposed method improves the \(5\)-way \(1\)-shot accuracy from \(73.49\%\) to \(76.07\%\), and 5-way 5-shot accuracy from \(86.64\%\) to \(91.16\%\).
On mini-ImageNet the proposed method improves the \(5\)-way \(1\)-shot accuracy from \(62.86\%\) to \(64.71\%\), and 5-way 5-shot accuracy from \(78.06\%\) to \(83.23\%\).


\section{Background}

\subsection{Few-shot classification}
\paragraph{Problem formulation}
In few-shot classification, The data set is divided into a base set, a validation set, and a novel set (the three sets disjoint from each other).
We are given sufficient training samples on the base set, and the objective is to solve a classification problem on the novel set, which only has few training samples.
The standard formulation is an \(N\)-way \(K\)-shot classification problem.
There are \(N\) classes sampled from the novel set, and \(K\) + \(Q\) non-repeating samples sampled for each class.
\(N\) * \(K\) samples form a support set for training the classifier, and \(N\) * \(Q\) samples form a query set for testing.
Generally, the value of \(K\) is selected from 1 or 5, and the value of \(N\) is selected from \(5\), \(10\), or \(20\).

\paragraph{Meta-learning based methods}
Meta-learning methods focus on learning how to learn or to quickly adapt to new information, which typically involve a meta-learner model that given a few training examples of a new task it tries to quickly learn a learner model that "solves" this new task.
Specifically,
Matching Nets~\citep{DBLP:conf/nips/VinyalsBLKW16} augment neural networks with external memories, make it possible to generate labels for unknown categories.
MAML~\citep{DBLP:conf/icml/FinnAL17} tries to get an excellent initial condition that the classifier can recognize novel classes with few labeled examples and a small number of gradient update steps.
LEO~\citep{DBLP:conf/iclr/RusuRSVPOH19} expands fast adaptation methods by learning a data-dependent latent generative representation of model parameters and performing gradient-based meta-learning in this low-dimensional latent space.
ProtoNet~\citep{DBLP:conf/nips/SnellSZ17} learns to classify examples by computing distances to prototype feature vectors.
RelationNet~\citep{DBLP:conf/cvpr/SungYZXTH18} replaces the Euclidean distance based non-parameter classifier used by ProtoNet with the CNN-based relation module.

\paragraph{Pre-training based methods}
Pre-training based methods generally uses a two-stage training strategy.
A feature extractor is obtained through the first stage of training, and the classifier used for few-shot classification tasks is obtained through the second stage of training.
SGM~\citep{DBLP:conf/iccv/HariharanG17} presents a way of “hallucinating” additional examples for novel classes.
In the first stage, train a feature generator that can generate new samples. In the second stage, a classifier is trained based on the original and generated samples.
DynamicFSL\citep{DBLP:conf/cvpr/GidarisK18} extends an object recognition system with an attention-based few-shot classification weight generator, which composes novel classification weight vectors by “looking” at a memory that contains the base classification weight vectors in the second stage.
Baseline++~\citep{DBLP:conf/iclr/ChenLKWH19} finds the advantages of the cosine-similarity based classifier, which generalizes significantly better features on novel categories than the dot-product based classifier.

\subsection{High-Resolution Net}
The ImageNet classification network, e.g., ResNet~\citep{DBLP:conf/cvpr/HeZRS16} adopts a high-to-low process aims to generate low-resolution and high-level representations.
Differently, High-Resolution Net (HRNet)~\citep{DBLP:conf/cvpr/0009XLW19}, which can maintain high-resolution representations through the whole process, conducts repeated multi-scale fusions by exchanging the information across the parallel multi-resolution subnetworks over and over through the whole process, performing extremely well in the pose estimation domain.

\section{Methodology}

Our work follows Pre-training based methods. 
However, we propose a novelty-prepared loss function, called self-compacting softmax loss (SSL), to replace traditional softmax loss (SL).
Further, we adopt a two-stage training strategy and introduce the High-Resolution Net as our backbone.
We present the details below.

\begin{figure}[t]
\centering
\subfigure[Before adjustment] {
\includegraphics[width=0.45\columnwidth]{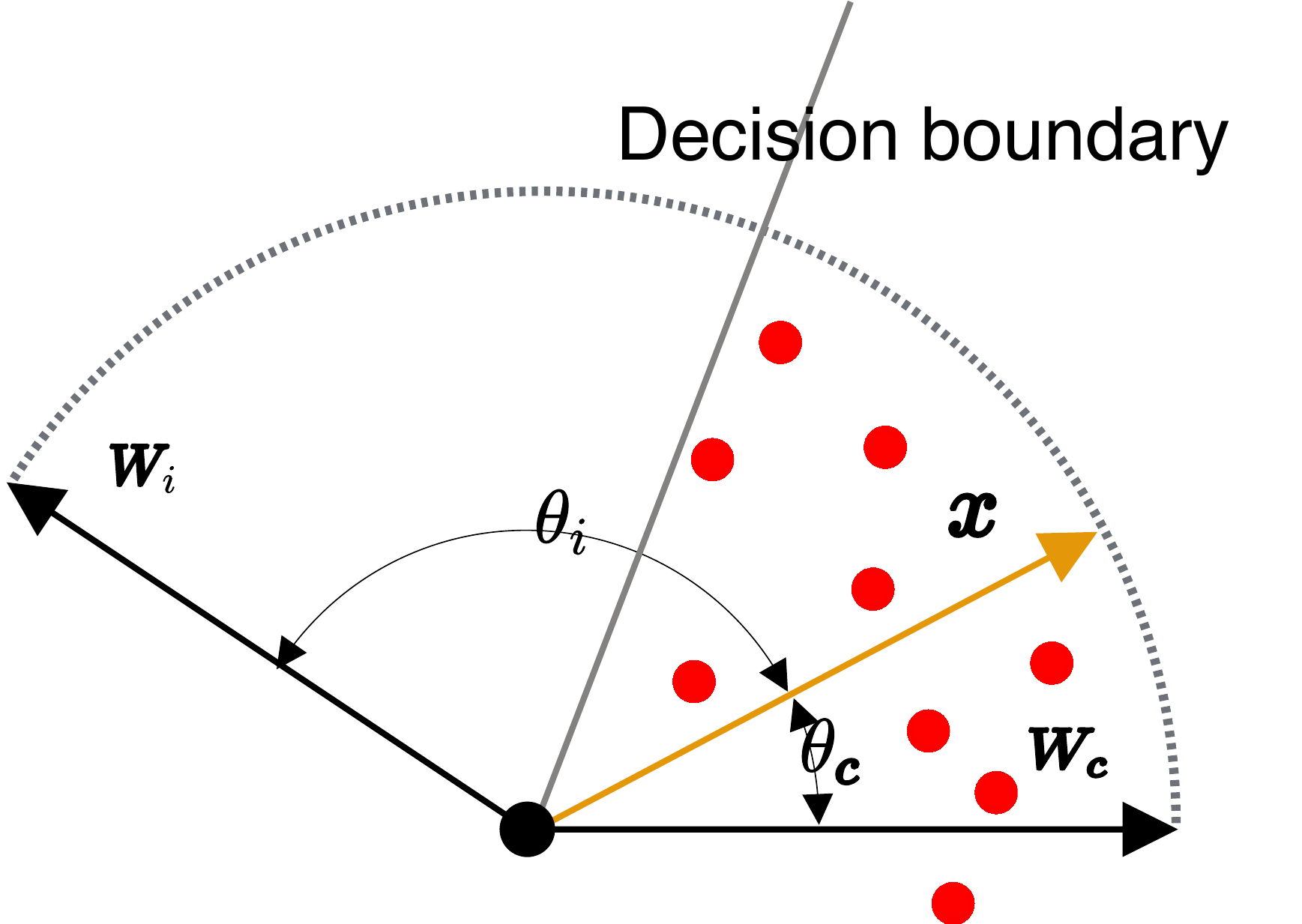}}
\subfigure[After adjustment] {
\includegraphics[width=0.45\columnwidth]{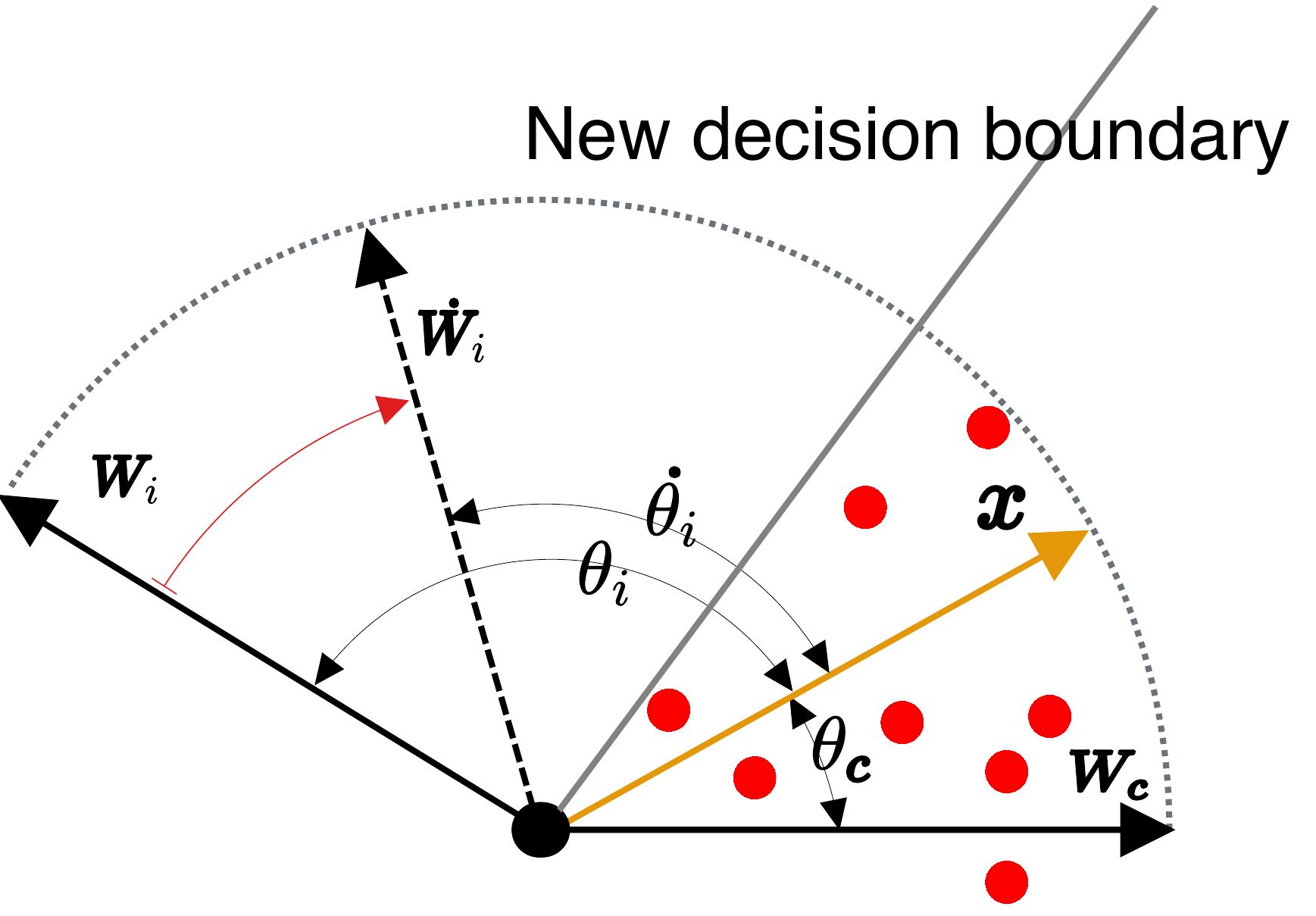}}
\caption{Example of adjusting the direction of a weight vector. The red dots represent the features of instances from class \(c\). The red arrow represents the adjustment of Eq.\ref{SSL_adjust}.
After the adjustment, \(w_i\) will be closer to \(w_c\).
And the new decision boundary (the angular bisector of \(w_c\) and \(w_i\)) will be closer to \(w_c\).
Therefore, when \(f_\theta\) is updated with the backpropagated gradient, the features of instances from class \(c\) will be closer to \(w_c\).}
\label{fig:adjustment}
\end{figure}

\subsection{Self-compacting softmax loss}
\label{section_SSL}
The pre-training process on the base set was commonly unaware of the future adaptation to other concept classes.
When using SL in this process, e.g., baseline++~\citep{DBLP:conf/iclr/ChenLKWH19}, in order to ensure the correct classification of the base set, SL will make the similarity between the features of different classes of the base set as low as possible.
Therefore, in the embedding space obtained from a classically fully trained feature extractor using SL, the base class feature clusters are far away from each other, almost entirely occupying the embedding space, and there is little embedding space for unseen classes, which keeps the feature extractor from well-fitting the novel classes.
In Section~\ref{Visualisation}, we will visualize this phenomenon to explain it.

We argue that one way to solve the above problem of using SL in few-shot classification tasks is to add a compact optimization objective while maintaining the classification optimization objective.
The classification optimization objective will make the feature clusters in the embedding space to be far away from each other, and the compact optimization objective will make the feature clusters in the embedding space to be close to each other.
When the above two optimization objectives are satisfied, the feature clusters in the embedding space will be as close as possible to each other while maintaining separability.
In this case, base classes take up less embedding space overall, and more embedding prepared space for novel classes.
Following this thought process, we propose a novelty-prepared loss function, called self-compacting softmax loss (SSL), which optimizes the compact objective in addition to the primary classification optimization objective.

We use a single fully connected layer to implement an SSL classifier \(C_{SSL}\). 
For an image feature embedding \(\phi(x)\) (\(x\in R^D,\phi(x)\in R^d\), the ground-truth of \(x\) is label \(c\)),
The mathematical form of the classifier \(C_{SSL}\) outputs the probability for each class is as follows:
\begin{eqnarray}
p_i = \frac{\exp(w_i^T \phi(x)+b_i)}
          {\sum_{j=1}^{C} \exp(w_j^T \phi(x)+b_j)}        
          \quad i\in C
\end{eqnarray}
Where \(w_i \in W\) and \(b_i \in B\), \(W\) denote the weights and \(B\) denote the biases of the FC layer.
It can be converted as:
\begin{eqnarray}
\label{cos_softmax_p}
p_i = \frac{\exp(\| w_i\| \|\phi(x)\| \cos\theta_i+b_i)}
      {\sum_{j=1}^{C} \exp(\| w_j\| \|\phi(x)\| \cos\theta_j+b_j)}
\end{eqnarray}
We remove the bias in the FC layer firstly and then normalize \(w\) and \(f_\theta(x)\) by L2-norm to let \(\|w\| =\|f_\theta(x)\|=1 \), to construct \(C_{SSL}\) as a cosine-similarity based classifier.
In this way, we decouple the magnitude of a weight tensor from its direction, forcing network learning to classify by differences in angular direction.
Cosine similarity scores \(S_{cos}\) is obtained by matrix multiplication of image feature vectors and weight vectors which represent categories.
\begin{eqnarray}
S_{cos}(w_i,\phi(x))=\alpha \|w_i\|\|\phi(x)\|
\quad w_i\in W
\end{eqnarray}
where \(\alpha\) is an amplification factor that helps with convergence.

For each instance \(x\), we adjust the direction of each weight vector \(w_i\) as Eq.\ref{SSL_adjust}.
\begin{eqnarray}
\label{SSL_adjust}
\dot{w_i} = \frac{|S_{cos}(w_i,\phi(x))-S_{cos}(w_c,\phi(x))|*w_c + w_i}
{\||S_{cos}(w_i,\phi(x))-S_{cos}(w_c,\phi(x))|*w_c + w_i \|}
\end{eqnarray}
where \(\dot{w_i}\) is the adjusted \(w_i\), \(w_i \in W\).

Therefore, \(C_{SSL}\) outputs the probability for each class is as follows:
\begin{eqnarray}
p_i = \frac{\exp(S_{cos}(\dot{w_i},\phi(x)))}{\sum_{i}^{C} \exp(S_{cos}(\dot{w_i},\phi(x)))}
\end{eqnarray}
and the SSL is as follows:\\
\begin{eqnarray}
SSL =-\log \frac{\exp(S_{cos}(w_c,\phi(x)))}{\sum_{i}^{C} \exp(S_{cos}(\dot{w_i
},\phi(x)))}
\end{eqnarray}

In Eq.\ref{cos_softmax_p}, the angular bisector of \(w_c\) and \(w_i\) forms a decision boundary.
After the adjustment of Eq.\ref{SSL_adjust}, each \(w_i\) (\(w_i\in W, i\not=c\)) will be closer to \(w_c\) to varying degrees according to the absolute value of the cosine similarity difference between \(\phi(x)\) and \(w_i\), \(\phi(x)\) and \(w_c\).
And the new decision boundary will be closer to \(w_c\).
When \(f_\theta\) is updated with the backpropagated gradient, the features of instances from class \(c\) will be closer to \(w_c\).
In Fig.\ref{fig:adjustment}, we show this process.

The adjustment of Eq.\ref{SSL_adjust} is the compact optimization objective, which requires the feature clusters in the embedding space to be close to each other.
It should be mentioned that the proximity of the feature clusters in the embedding space is proportional to the angle between the prototypes (\(w_i \in W\), \(w_i\) is the prototype (class center) of \(i\)th class).
Therefore, unlike SL, which eventually makes the angles between the prototypes of different base classes obtuse, the SSL eventually makes the angles between the prototypes of different base classes tend to orthogonal (In Section~\ref{Ablation study and analysis}, we will explain it by experiments).
In this way, the separability between the base classes is maintained, more space is prepared for the novel classes and greatly improved the generalization on the novel set.
In Section~\ref{Visualisation}, we show the Visualization of features obtained using the SSL and using SL by t-SNE, which also proves the superior generalization performance brought by SSL.

{}
\subsection{Training framework}
\begin{figure}[t]
\begin{minipage}[b]{1.0\linewidth}
  \centering
  \centerline{\epsfig{figure=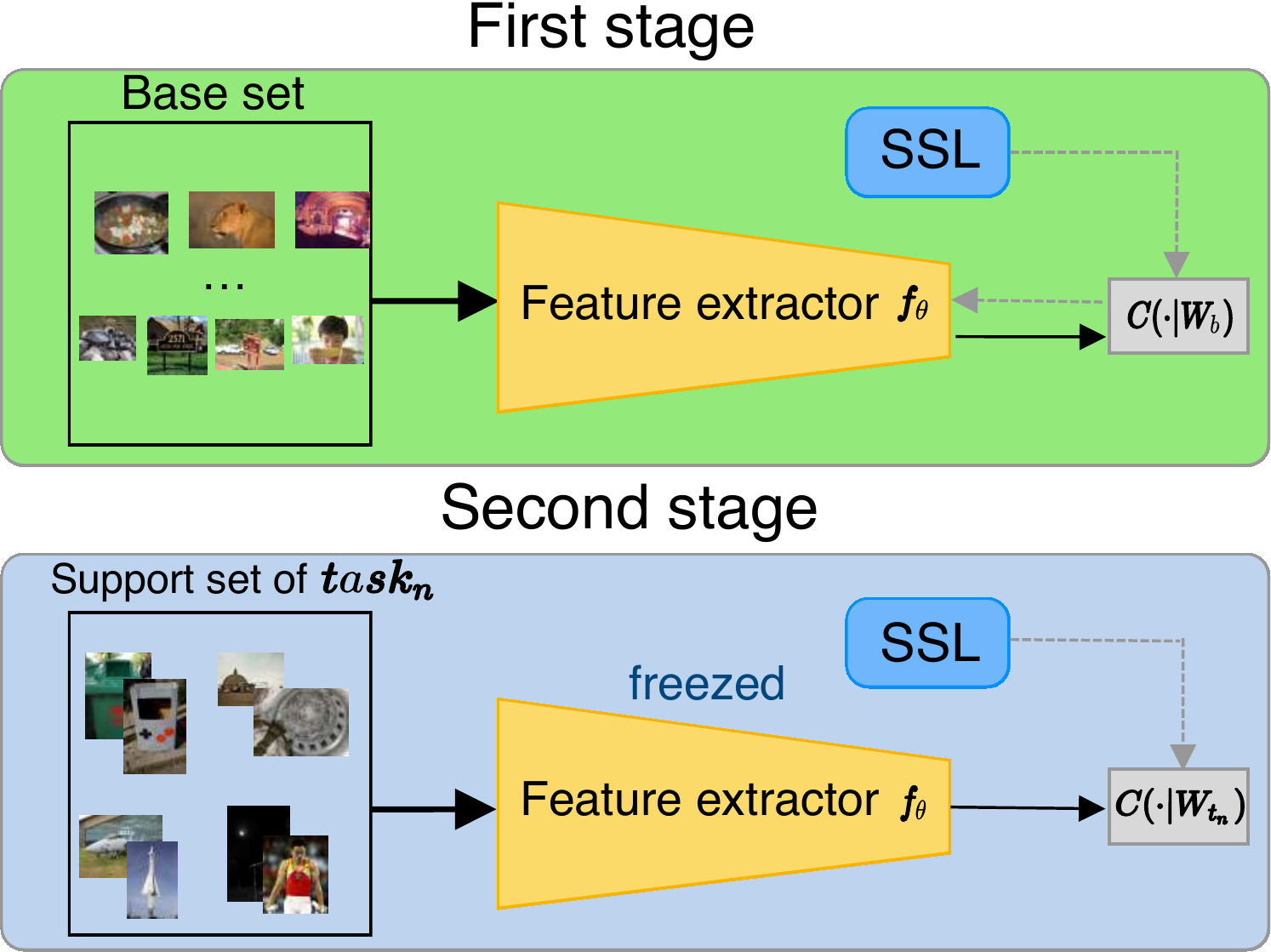,width=8cm,height=6cm}}
\caption{Overview of the two-stage training strategy.}
\label{fig:Overview}
\end{minipage}
\end{figure}

We adopt a two-stage training strategy.
At the first stage, we train a feature extractor \(f_{\theta}\) (parametrized by the network parameters \(\theta\)) followed by a classifier \(C(\cdot | W_b)\) by minimizing the SSL on the whole base set.
\(C(\cdot | W_b)\) parametrized by the weight matrix \(W_b \in R^{d \times c_b}\), \(W_b = [w_1,\cdot\cdot\cdot,w_{c_b}]\).
\(d\) is the dimension of the feature and \(c_b\) is the number of categories of the base set.
When testing on the \(N\)-way \(K\)-shot tasks, is the second stage.
We freeze the \(f_{\theta}\), use the support set to train a task-specific classifier \(C(\cdot | W_t)\) for each task that sampled from the novel set.
\(C(\cdot | W_t)\) parametrized by the weight matrix \(W_t \in R^{d \times c_t}\), \(c_t\) is the number of categories of the task.
Same as at the first stage, using the SSL.
As mentioned in Section~\ref{section_SSL}, we construct our \(C(\cdot | W_b)\) and \(C(\cdot | W_t)\) as cosine-similarity based classifiers.
In cosine-similarity based classifiers, \(w_i\) (\(w_i \in W\)) is the prototype vector (class center vector) of \(i\)th class.
Fig.\ref{fig:Overview} is an overview of our two-stage training strategy.
In Section~\ref{Ablation study and analysis}, we will analyze in detail the different effects of using the SSL and SL in each stage.

Besides, we argue that the high-resolution feature is a useful complement to few-shot classification tasks.
However, most existing methods pass the input through a deep network, typically consisting of high-to-low resolution sub-network blocks that are connected in series, getting a low-resolution representation finally.
The massive loss in resolution wastes some useful information, which is particularly bad for few-shot classification in which labeled examples are not abundant.
So we use HRNet in our few-shot classification tasks as \(f_\theta\) to get high-resolution features fusion by multi-scale features, which further improves the performance of our method.
For a fair comparison with several state-of-the-art baselines, we use HRNet-W18-C-Small-v1~\citep{DBLP:journals/corr/abs-1908-07919}, of which parameters (15.6M) and GFLOPs are similar to ResNet-18.

\section{Experiment}
We evaluate our method on two few-shot classification benchmark: CUB-200-2011~\citep{wah2011caltech} and mini-ImageNet~\citep{DBLP:conf/nips/VinyalsBLKW16}, and compare against several state-of-the-art baselines including Matching Nets~\citep{DBLP:conf/nips/VinyalsBLKW16},  MAML~\citep{DBLP:conf/icml/FinnAL17}, ProtoNet~\citep{DBLP:conf/nips/SnellSZ17}, Relation Net~\citep{DBLP:conf/cvpr/SungYZXTH18}, FEAT~\citep{DBLP:journals/corr/abs-1812-03664}, DCN~\citep{DBLP:journals/corr/abs-1811-07100}, DynamicFSL~\citep{DBLP:conf/cvpr/GidarisK18}, TADAM~\citep{DBLP:conf/nips/OreshkinLL18}, LEO~\citep{DBLP:conf/iclr/RusuRSVPOH19}, SCA~\citep{DBLP:journals/corr/abs-1905-10295}, Baseline++~\citep{DBLP:conf/iclr/ChenLKWH19}.

\subsection{Dataset}
CUB-200-2011(CUB) dataset contains 200 classes and 11,788 images in total. Following the evaluation protocol of~\citep{hilliard2018few}, we randomly split the dataset into 100 base, 50 validation, and 50 novel classes. The mini-ImageNet dataset consists of a subset of 100 classes from the ImageNet~\citep{DBLP:conf/cvpr/DengDSLL009} dataset and contains 600 images for each class. We use the follow-up setting provided by~\citep{DBLP:conf/iclr/RaviL17}, which is composed of randomly selected 64 base, 16 validation, and 20 novel classes.

\subsection{Implementation details}
In the first stage training for the agnostic feature extractor \(f_\theta\), we
train 200 epochs with a batch size of 200.
The SGD optimizer is employed, and the initial learning rate is set to be 0.1. Then we adjust the learning rate by 0.1 every 50 epochs.
In the second stage, we average the results over 600 tasks.
In each task, we randomly sample \(N\) classes from novel classes, and in each class, we also pick \(K\) instances for the support set and 16 instances for the query set.
For each task, we use the entire support set to train a task-specific classifier for 100 iterations with a batch size of 4, and the learning rate is 0.01.
The SGD optimizer is employed.
We apply standard data augmentation, including random crop, left-right flip, and color jitter in both two stages.

In the first stage, ProtoNet, RelationNet, and Baseline++, they only use the base set to train and the validation set to verify.
However, in other works such as DCN, FEAT, as per common practice, they use the base set and the validation set together to train.
To be fair, we provide the result of only using the base set to train and the result of using the base set and the validation set together to train.
* distinguishes the latter in Table~\ref{tab:table1}.


\begin{table*}[t]
\scriptsize
\caption{CUB and mini-ImageNet \(5\)-way Acc. The results with * use both base set and validation set to train. The results with $\dag$ reported by~\citep{DBLP:conf/iclr/ChenLKWH19}. The backbone with $\ddagger$ indicates that in addition to the feature extractor of the network structure mentioned, additional network components are used.}
\label{tab:table1}
\setlength{\tabcolsep}{1.mm}
\centering
\begin{tabular}{llllll}
\toprule
\multirow{2}{*}{Model} &\multirow{2}{*}{Backbone}& \multicolumn{2}{c}{CUB} & \multicolumn{2}{c}{mini-ImageNet} \\
                       &                         & \multicolumn{1}{c}{1-shot} & \multicolumn{1}{c}{5-shot}& \multicolumn{1}{c}{1-shot} & \multicolumn{1}{c}{5-shot} \\
\midrule
MatchingNet~\citep{DBLP:conf/nips/VinyalsBLKW16}\dag   &ResNet-18   &\(73.49\pm0.89\%\)     &\(84.45\pm0.58\%\)  &\(52.91\pm0.88\%\)    &\(68.88\pm0.69\%\)\\
MAML~\citep{DBLP:conf/icml/FinnAL17}\dag        &ResNet-18   &\(68.42\pm1.07\%\)     &\(83.47\pm0.62\%\)     &\(49.61\pm0.92\%\)       &\(65.72\pm0.77\%\)\\
ProtoNet~\citep{DBLP:conf/nips/SnellSZ17}\dag    &ResNet-18   &\(72.99\pm0.88\%\)&\(86.64\pm0.51\%\)&\(54.16\pm0.82\%\)       &\(73.68\pm0.65\%\)\\
RelationNet~\citep{DBLP:conf/cvpr/SungYZXTH18}\dag      &ResNet-18    &\(68.58\pm0.94\%\)     &\(84.05\pm0.56\%\)      &\(52.48\pm0.86\%\)          &\(69.83\pm0.68\%\)\\
FEAT*~\citep{DBLP:journals/corr/abs-1812-03664}             &ResNet$\ddagger$      &\(68.65\pm0.22\%\)     &\(83.03\pm0.15\%\)      &\(62.60\pm0.20 \%\)          &\(78.06\pm0.15\%\)\\
DCN*~\citep{DBLP:journals/corr/abs-1811-07100}               &SENet$\ddagger$      & ---                   &---                     &\(62.88\pm0.83\%\)          &\(75.84\pm0.65\%\)\\ 
DynamicFSL~\citep{DBLP:conf/cvpr/GidarisK18}             &ResNet-10$\ddagger$      &\(56.20\pm0.86\%\)     &\(73.00\pm0.64\%\)      &\(56.20\pm0.86\%\)          &\(73.00\pm0.64\%\)\\
TADAM~\citep{DBLP:conf/nips/OreshkinLL18}      &ResNet-12$\ddagger$                                                   &---                    &---                     &\(58.5\pm0.3 \%\)           &\(76.7\pm0.3 \%\)\\
LEO ~\citep{DBLP:conf/iclr/RusuRSVPOH19}            & WRN-28-10$\ddagger$                   &---                    &---                     &\(61.76\pm0.08\%\)          &\(77.59\pm0.12\%\)\\
SCA*\citep{DBLP:journals/corr/abs-1905-10295}        &DenseNet$\ddagger$  &\(70.46\pm1.18\%\)            &\(85.63\pm0.66\%\)       &\(62.86\pm0.79\%\)&\(77.64\pm0.40\%\)\\
Baseline++~\citep{DBLP:conf/iclr/ChenLKWH19}         &ResNet-18       &\(67.02\pm0.90\%\)     &\(83.58\pm0.54\%\)      &\(51.87\pm0.77\%\)          &\(75.68\pm0.63\%\)\\
\midrule
SSL    &ResNet-18    &\(69.14\pm0.87\%\)     &\(86.96\pm0.52\%\)      &\(60.14\pm0.80\%\)          &\(79.98\pm0.57\%\)\\
SSL    &HRNet         &\(70.36\pm 0.86\%\)    &\(87.92\pm 0.46\%\)     &\(61.97\pm0.88\%\)          &\(81.95\pm0.57\%\)\\
SSL*   &ResNet-18   &\(74.05\pm 0.83\%\)    &\(89.92\pm 0.41\%\)     &\(60.98\pm0.81\%\)          &\(80.61\pm 0.49\%\)\\
SSL*   &HRNet      &\(\bm{76.07\pm0.82\%}\)&\(\bm{91.16\pm0.37\%}\) &\(\bm{64.71\pm0.83\%}\)     &\(\bm{83.23\pm0.54\%}\)\\
\bottomrule
\end{tabular}
\end{table*}

\subsection{Evaluation}
We report the mean of 600 randomly generated test episodes as well as the 95\% confidence intervals.
In Table~\ref{tab:table1}, we report the \(5\)-way \(1\)-shot, \(5\)-way \(5\)-shot results on CUB and mini-ImageNet, and compare them with the state-of-the-art baselines mentioned above. 
In Table~\ref{tab:table2}, we report the \(N\)-way \(5\)-shot results of our approach on mini-ImageNet.
In Table~\ref{tab:table3}, we report \(5\)-way \(5\)-shot accuracy under the cross-domain scenario.
We use mini-ImageNet as our base class and the 50 validation and 50 novel class from CUB as~\citep{DBLP:conf/iclr/ChenLKWH19}.
Evaluating the cross-domain scenario allows us to illustrate the performance of our method when domain shifts are present.

\begin{table}[t]
\centering\small
\setlength{\tabcolsep}{2.2mm}
\caption{mini-ImageNet \(N\)-way \(5\)-shot Acc. The results with {\dag} reported by~\citep{DBLP:conf/iclr/ChenLKWH19}.}
\label{tab:table2}
\begin{tabular}{lll}
\toprule
N-way test         & 10-way                  & 20-way         \\
\midrule
MatchingNet\dag        &\(52.27\pm0.46\%\)       &\(36.78\pm0.25\%\)\\
ProtoNets\dag          &\(59.22\pm0.44\%\)       &\(44.96\pm0.26\%\)\\
RelationNet \dag       &\(53.88\pm0.48\%\)       &\(39.17\pm0.25\%\)\\
Baseline++\dag         &\(63.40\pm0.44\%\)       &\(50.85\pm0.25\%\)\\
\midrule
SSL (ResNet-18)    &\(65.11\pm0.45\%\)       &\(52.99\pm0.25\%\) \\
SSL (HRNet)       &\(\bm{67.85\pm0.41\%}\)  &\(\bm{56.15\pm0.25\%}\)\\
\bottomrule
\end{tabular}
\end{table}

\begin{table}[t]
\centering\small
\caption{\(5\)-way \(5\)-shot accuracy under the cross-domain scenario. The results with $\dag$ reported by~\citep{DBLP:conf/iclr/ChenLKWH19}.}
\label{tab:table3}
\setlength{\tabcolsep}{4.5mm}
\begin{tabular}{ll}
\toprule
                       & mini-ImageNet\(\rightarrow\) CUB         \\
\midrule
MatchingNet\dag        &\(53.07\pm0.74\%\)\\
ProtoNets\dag          &\(62.02\pm0.70\%\)\\
MAML\dag               &\(51.34\pm0.72\%\)\\
RelationNet \dag       &\(57.71\pm0.73\%\)\\
Baseline++\dag         &\(62.04\pm0.76\%\)\\
\midrule
SSL (ResNet-18)        &\(67.20\pm0.71\%\)\\
SSL (HRNet)            &\(\bm{69.49\pm0.70\%}\)\\
\bottomrule
\end{tabular}
\end{table}

\begin{figure*}[h]
\centering
\subfigure[base (SL) and base (SSL).]{
\begin{minipage}[t]{0.3\linewidth}
\centering
\includegraphics[width=1in]{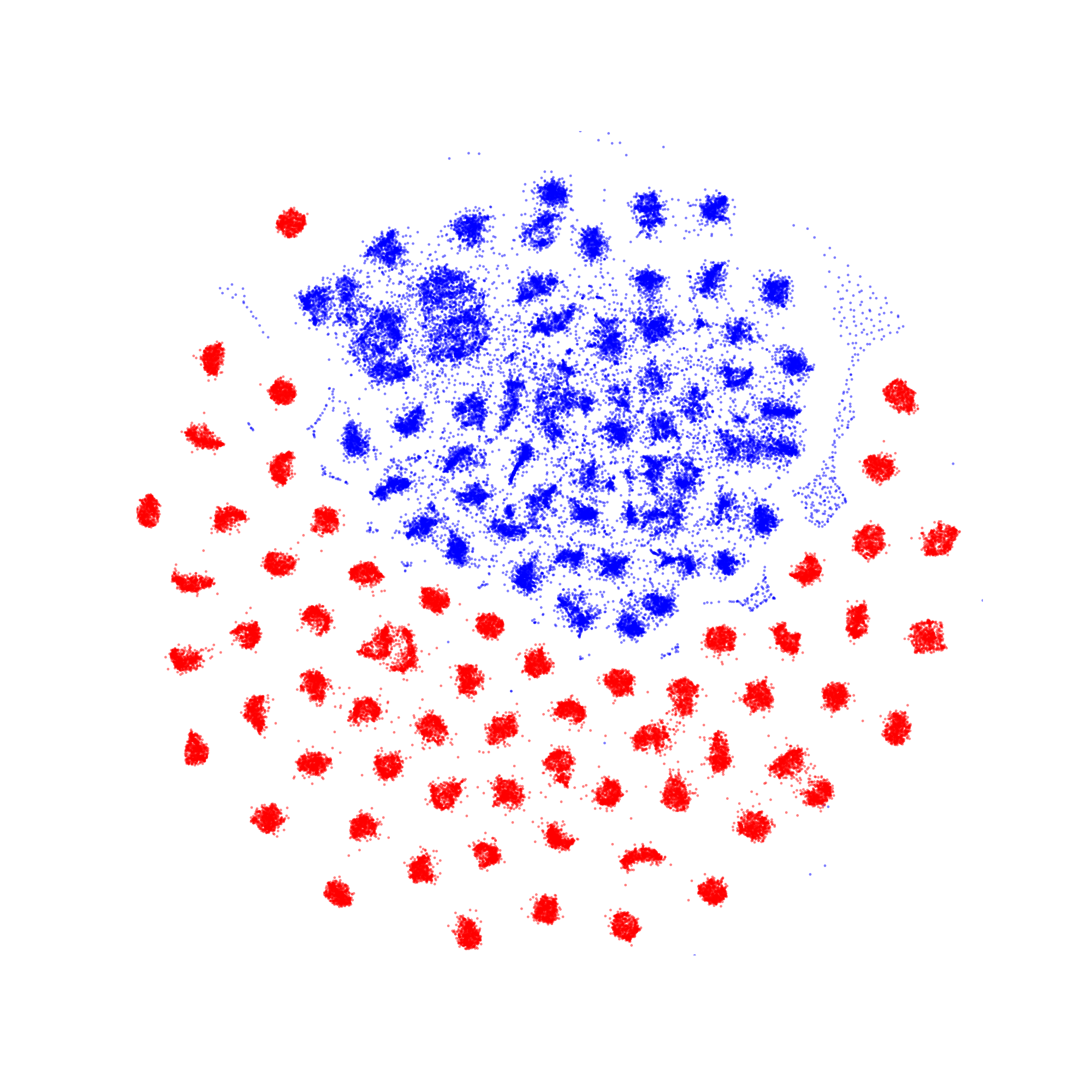}
\label{fig:t-sne:base+base}
\end{minipage}%
}
\subfigure[base and novel (SL).]{
\begin{minipage}[t]{0.3\linewidth}
\centering
\includegraphics[width=1.5in]{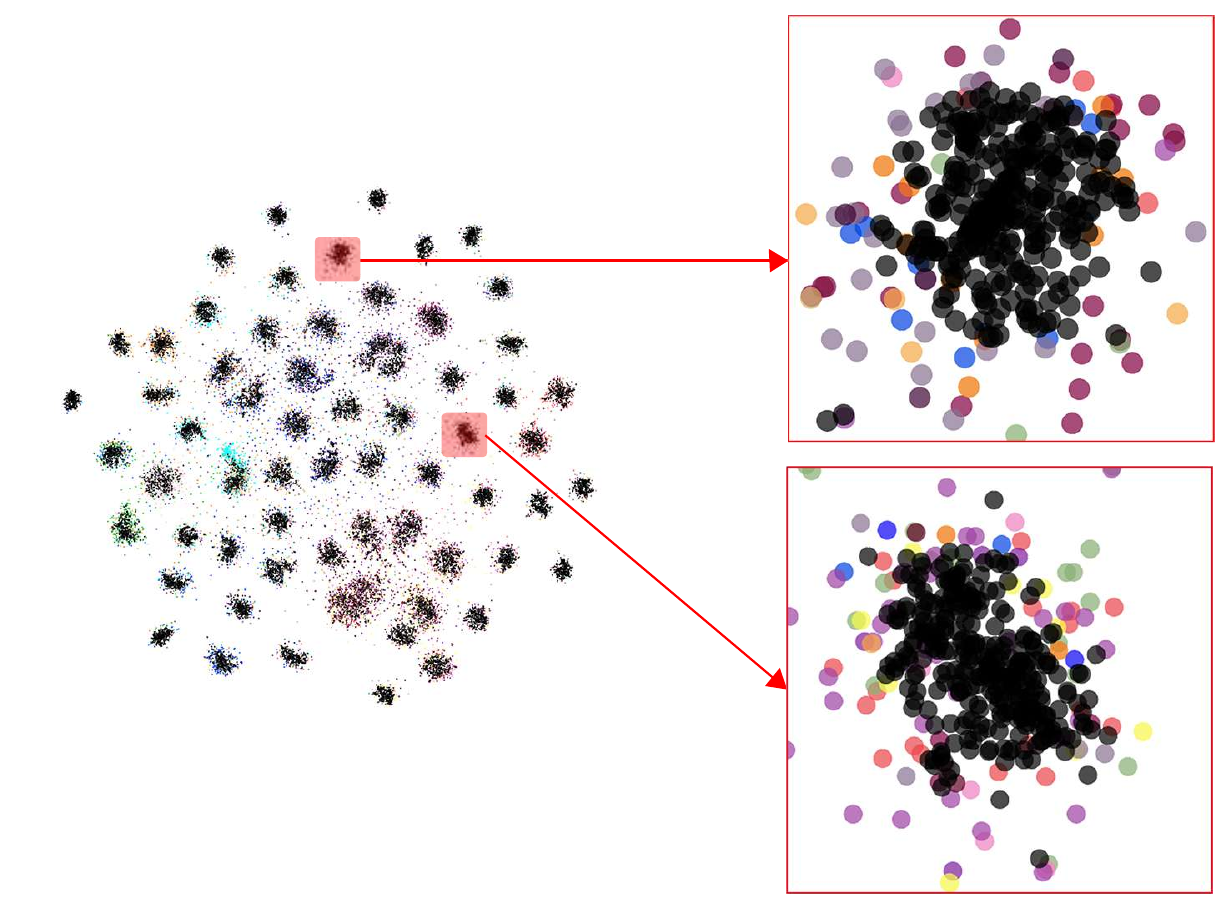}
\label{fig:t-sne:base+novel-noSSL}
\end{minipage}%
}
\subfigure[base and novel (SSL).]{
\begin{minipage}[t]{0.3\linewidth}
\centering
\includegraphics[width=1.5in]{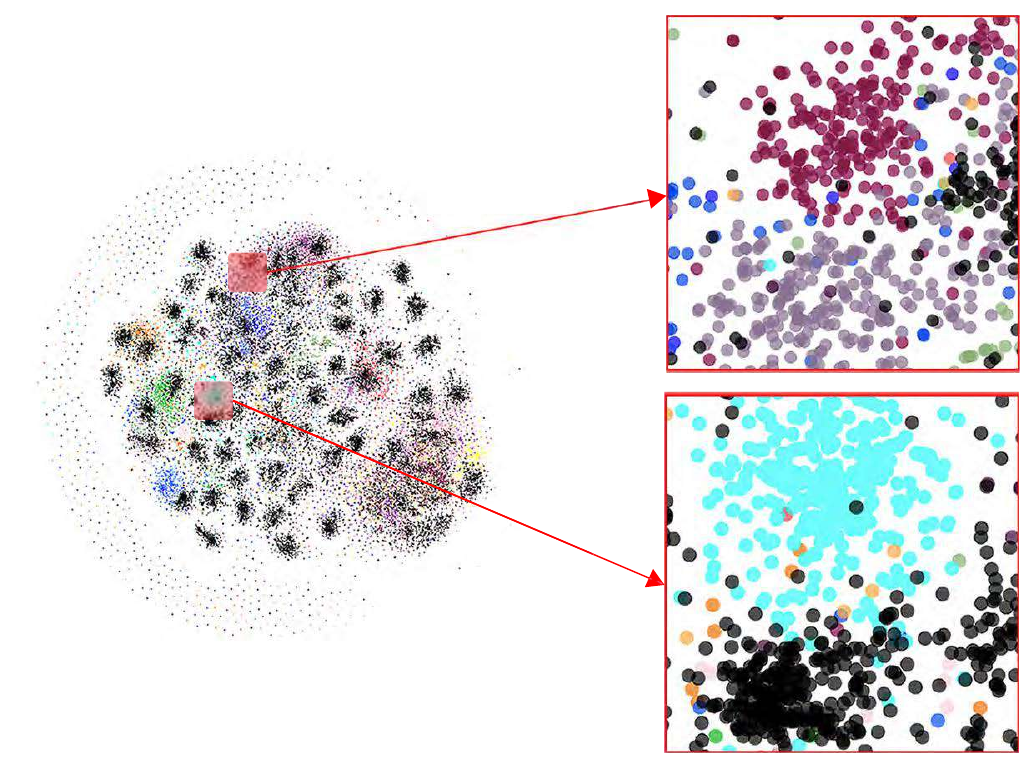}
\label{fig:t-sne:base+novel-SSL}
\end{minipage}%
}
\caption{
Visualization of features by t-SNE.
The data set used mini-ImageNet.
In (a), the blue dots represent the base set features obtained by a feature extractor trained with SSL, and the red dots represent the base set features obtained by a feature extractor trained with SL (softmax loss).
In (b) and (c), we plotted the relationship between the base set features and the novel set features.
The black dots represent the base set features, the other colored dots represent the novel set features, and different colors represent different novel classes.
The features in (b) obtained by a feature extractor trained with SL and the features in (c) obtained by a feature extractor trained with SSL.
It can be seen from the figure that when using SL, the base set features almost fill the entire embedding space, and the novel set features of the same class scatter around feature clusters of different classes of the base set and cannot converge into clusters. In contrast, when using SSL, the base set features are more compact and occupy less embedding space, and the novel set features mainly scatter among feature clusters of different classes of the base set and converge into clusters.
Best viewed in colors.
}
\label{fig:t-sne}
\end{figure*}

\subsection{Visualisation}
\label{Visualisation}

To illustrate the previous point, that is, in the embedding space obtained by the feature extractor fully trained using SL, the base set features almost fill the entire embedding space, so there is little available embedding space for the novel set features, while SSL can prevent the full occupancy of the embedding space, so there is more space prepared for the novel set features.
In Fig.\ref{fig:t-sne}, we show the Visualization of features obtained using the SSL and using SL by t-SNE.
The data set used mini-ImageNet.

In Fig.\ref{fig:t-sne:base+base},, it can be seen that the base set features ob- tained using the SSL (blue dots) occupy less space than the base set features obtained using the SL (red dots).
In Fig.\ref{fig:t-sne:base+novel-noSSL}, when using SL, the novel set features of the same class scatter around feature clusters of different classes of the base set and cannot converge into clusters.
In contrast, in Fig.\ref{fig:t-sne:base+novel-SSL}, when using the SSL, the novel set features mainly scatter among feature clusters of different classes of the base set and converge into clusters.

\begin{figure}[t]
\begin{minipage}[b]{1.0\linewidth}
  \centering
  \centerline{\epsfig{figure=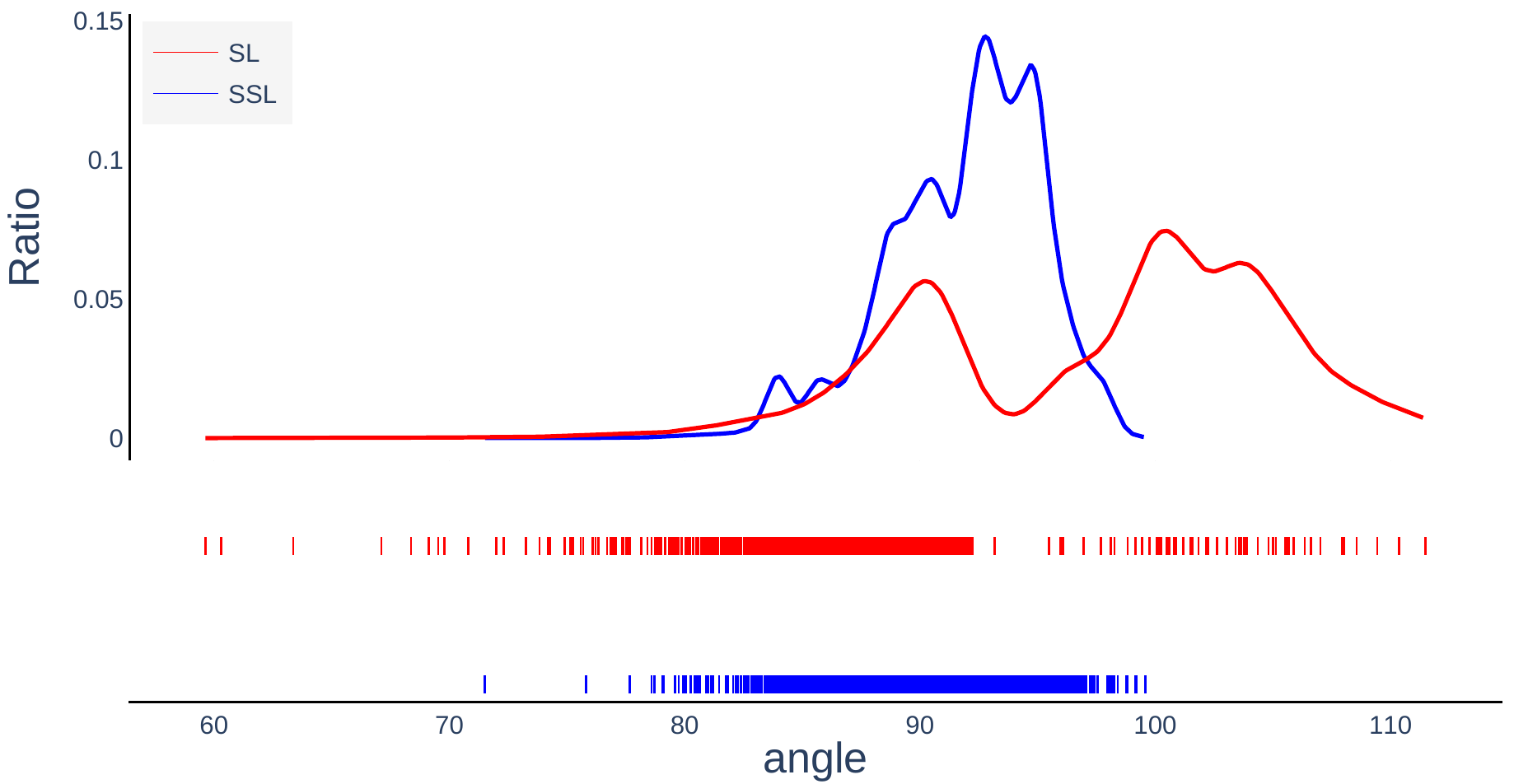,width=8.cm,height=4.5cm}}
\caption{Distribution of angles between prototypes. Red represents the distribution of angles between prototypes obtained using SL(softmax loss), and blue represents the distribution of angles between prototypes obtained using SSL.}
\label{fig:prototype_angles}
\end{minipage}
\end{figure}

\begin{figure}[t]
\begin{minipage}[b]{1.0\linewidth}
  \centering
  \centerline{\epsfig{figure=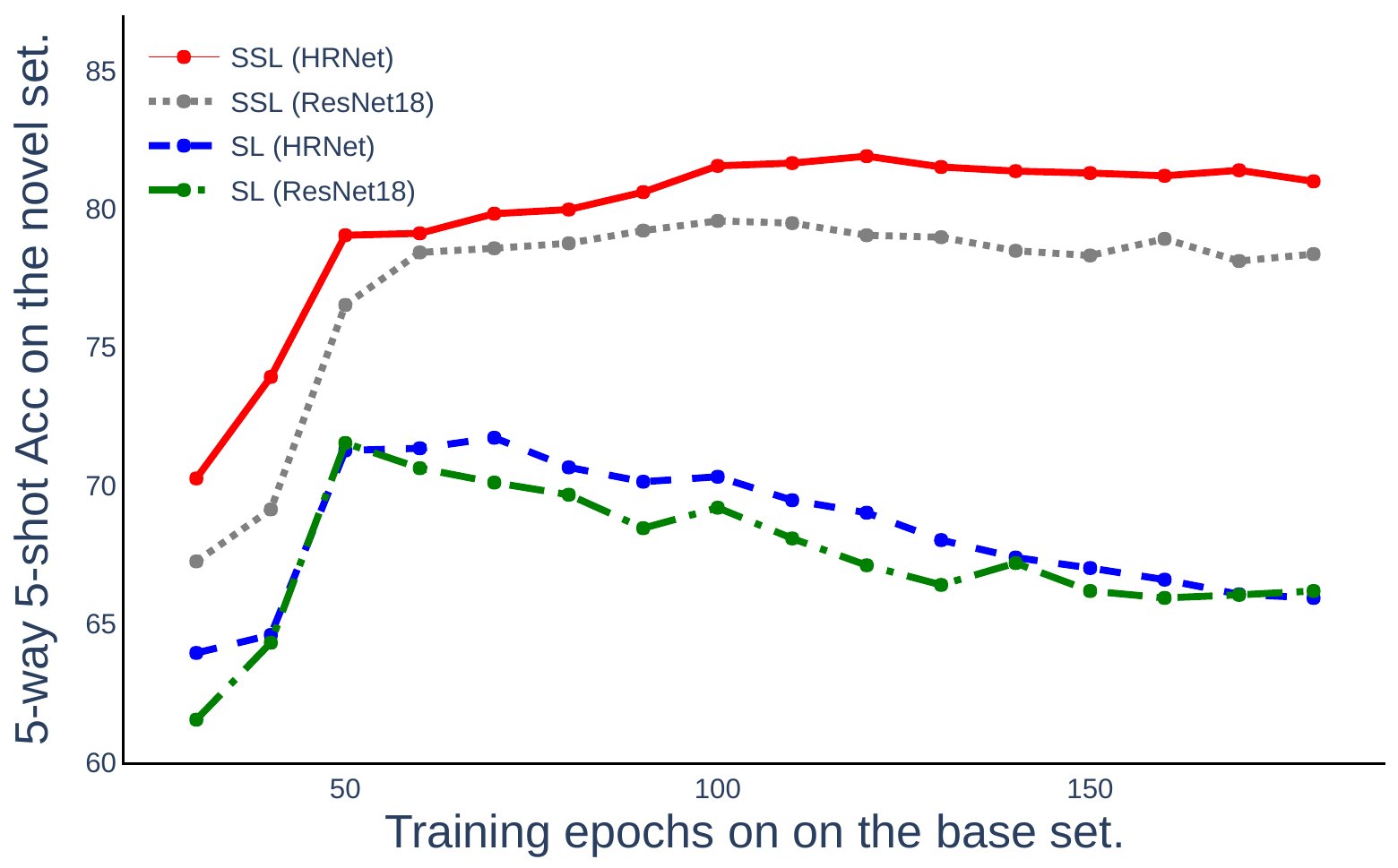,width=8.cm,height=5.0cm}}
\caption{SSL vs SL(softmax loss). mini-ImageNet \(5\)-way \(5\)-shot Acc.}
\label{fig:SSLVSL}
\end{minipage}
\end{figure}

\subsection{Ablation study and analysis}
\label{Ablation study and analysis}
In this section, we perform further analysis for our approach on the mini-ImageNet.

\paragraph{What is the difference between the features obtained using SL and the features obtained using the SSL?}
To answer this question, we extract weight matrix \(W_b = [w_1,\cdot\cdot\cdot,w_{c_b}]\) from \({C(\cdot | W_b)}_{SL}\)
and weight matrix \( \dot{W_{b}} = [ \dot{w_1},\cdot\cdot\cdot, \dot{w_{c_b}}]\) from \({C(\cdot |  \dot{W_b})}_{SSL}\).
\(C(\cdot | W_b)\) and \(C(\cdot | \dot{W_b})_{SSL}\) are both trained on mini-ImageNet).
As mentioned in section 3.2, \(w_i\) is the prototype (class center) of \(i\)th class in the base set.
We explain the characteristics of features through prototypes.
In Fig.\ref{fig:prototype_angles}, we show the distribution of angles between prototypes using SL and the distribution of angles between prototypes using SSL.
Obviously, compared with SL, the SSL makes the prototypes of the base set tend to be orthogonal.
In this way, the separability between the base classes is maintained (mini-ImageNet 5-way 5-shot accuracy of the base set from \(99.43\pm0.07\%\) to \(95.04\pm0.30\%\)), more space is prepared for the novel classes and greatly improved the generalization on the novel set (mini-ImageNet 5-way 5-shot accuracy of the novel set from \(71.74\pm0.69\%\) to \(81.95\pm0.57\%\)).

\paragraph{Effects of using SSL at each stage.}
How much difference between using the SSL and using SL in each stage?
To answer this question, we report the result of using the SSL in both two stages, using the SSL only in the first stage, using the SSL only in the second stage, and using SL in both two stages instead of the SSL. Results in Table~\ref{tab:table4} clearly show that the SSL improves performance both in two stages.

\paragraph{The influence of the high-resolution feature.}
We notice that the high-resolution feature has brought a considerable improvement (mini-ImageNet \(5\)-way \(5\)-shot accuracy of the novel set from~\(79.98\pm0.57\%\) to~\(81.95\pm0.57\%\) ) in our approach, which has led us to question what results can be obtained if only the high-resolution features are used?
In Fig.\ref{fig:SSLVSL}, we demonstrate the result of using HRNet trained with SSL, using HRNet trained with SL, using ResNet-18 trained with SSL, and using ResNet-18 trained with SL.
We can see from the figure that when SL is used, the results of ResNet-18 and HRNet are very close (\(71.55\%\),~\(71.74\%\) respectively), and both have obvious overfitting phenomena.
In contrast, when SSL is used, both of them get better results, and the curves eventually converge smoothly.
At this time, compared to ResNet-18, HRNet has brought a significant improvement.
Therefore, we conclude that although the high-resolution features are a useful complement to few-shot classifications, using it alone is not enough to solve the problem.
SL is not sufficient to drive the network to learn useful high-resolution features under limited labeled data.
Consequently, we think the SSL can prevent the full occupancy of the embedding space, so the generalization performance of the embedding space obtained by \(f_\theta\) better, which is the key for us to obtain such superior performance.

\paragraph{Impact of input size on performance.}
We find the results of Matching Nets, MAML, ProtoNet, RelationNet, Baseline++ reported in~\citep{DBLP:conf/iclr/ChenLKWH19} and DCN use input size of \(224\times 224\), but LEO uses input size of \(80\times 80\), FEAT, DynamicFSL and TADAM use input size of \(84\times 84\).
Therefore, to analyze the impact of input size on performance, we report our results under the input size of \(224\times 224\) and \(84\times 84\) in Table~\ref{tab:table5}.
Our method maintains good performance at low resolutions.
Satisfactory, HRNet performs extremely well in low-resolution compared to ResNet-18.
This also fully proves that the high-resolution feature are a useful complement to few-shot classifications.

\begin{table}[t]
\centering\small
\caption{Ablation analysis on using the SSL or SL (softmax loss) in each stage. mini-ImageNet \(5\)-way \(5\)-shot Acc.}
\label{tab:table4}
\begin{tabular}{llll}
\toprule
Stage1                &Stage2                      & ResNet-18                     & HRNet    \\
\midrule
SSL                   &SSL                        & \( 79.98\pm0.57\%\)          &\(81.95\pm0.57\%\)      \\
SSL                   &SL                         &\(79.01 \pm 0.54\%\)          &\(80.75\pm0.59\%\)      \\
SL                    &SSL                         &\(70.12\pm0.61\%\)            &\( 72.37\pm0.64\%\)     \\
SL                    &SL                         & \(69.09\pm0.70\%\)           &\( 71.74\pm0.69\%\)     \\
\bottomrule
\end{tabular}
\end{table}

\begin{table}[t]
\centering\small
\caption{Impact of input size on performance. mini-ImageNet \(5\)-way \(5\)-shot Acc.}
\label{tab:table5}
\setlength{\tabcolsep}{4.5mm}
\begin{tabular}{lll}
\toprule
Input size              & ResNet-18                     &HRNet \\
\midrule
\(224\times 224\)       &\(79.98 \pm 0.57\%\)          &\(81.95\pm0.57\%\) \\
\(84\times 84\)         &\(74.02\pm0.65\)\%            &\(79.01\pm0.61\%\) \\
\bottomrule
\end{tabular}
\end{table}

\section{Conclusions}

In this paper, we propose a novelty-prepared loss function designed for cosine-similarity based classifiers, which can prevent the full occupancy of the embedding space.
Thus the model is more prepared to learn new classes.
Moreover, we use the high-resolution features to retain more useful information.
Our method is based on pre-training based methods, with a novel perspective and simple implementation, which surpasses prior state-of-the-art approaches on CUB-200-2011 and mini-ImageNet dataset.

\bibliographystyle{plainnat}
\bibliography{ijcai20}

\end{document}